\pdfoutput=1
\documentclass[11pt]{article}

\usepackage[]{acl}

\usepackage{times}
\usepackage{latexsym}

\usepackage[T1]{fontenc}
\usepackage{enumitem}
\usepackage{amssymb}
\usepackage{amsmath}
\usepackage{graphicx}
\usepackage{subfig}
\usepackage[most]{tcolorbox}
\definecolor{softblue}{RGB}{142,201,253}
\definecolor{softorange}{RGB}{252,180,117}
\definecolor{satblue}{RGB}{67,165,252}
\definecolor{satorange}{RGB}{250,139,42}

\newtcbox{\entityOneFull}{on line,colback=softblue!25,colframe=softblue,size=fbox,arc=7pt, box align=base, before upper=\strut, top=-3pt, bottom=-3pt, boxrule=1.5pt}
\newtcbox{\entityOneTitle}{on line,colback=satblue,size=fbox,arc=3pt, box align=base, before upper=\strut, boxrule=0pt, fontupper=\color{white}, top=-2pt, bottom=-3pt,}
\newtcbox{\entityTwoFull}{on line,colback=softorange!25,colframe=softorange,size=fbox,arc=7pt, box align=base, before upper=\strut, top=-3pt, bottom=-3pt, boxrule=1.5pt}
\newtcbox{\entityTwoTitle}{on line,colback=satorange,size=fbox,arc=3pt, box align=base, before upper=\strut, boxrule=0pt, fontupper=\color{white}, top=-1.5pt, bottom=-2.5pt,}

\usepackage{tikz}
\newcommand*{\circled}[2][]{\tikz[baseline=(C.base)]{
    \node[inner sep=0pt] (C) {\vphantom{1g}#2};
    \node[draw, circle, inner sep=1pt, yshift=0pt] 
        at (C.center) {\vphantom{1g}};}}

\usepackage[utf8]{inputenc}

\usepackage{microtype}

\usepackage{inconsolata}

\title{Efficient Information Extraction in Few-Shot Relation Classification through Contrastive Representation Learning}

\author{
	Philipp Borchert$^{1,2}$, Jochen De Weerdt$^2$, Marie-Francine Moens$^3$\\
	\textsuperscript{1}IESEG School of Management, 3 Rue de la Digue, 59000 Lille, France\\
	\textsuperscript{2}Research Centre for Information Systems Engineering, KU Leuven, Belgium\\
	\textsuperscript{3}Department of Computer Science, KU Leuven, Belgium\\
}

\begin{document}
\maketitle
\begin{abstract}

Differentiating relationships between entity pairs with limited labeled instances poses a significant challenge in few-shot relation classification. Representations of textual data extract rich information spanning the domain, entities, and relations. In this paper, we introduce a novel approach to enhance information extraction combining multiple sentence representations and contrastive learning. While representations in relation classification are commonly extracted using entity marker tokens, we argue that substantial information within the internal model representations remains untapped. To address this, we propose aligning multiple sentence representations, such as the \texttt{[CLS]} token, the \texttt{[MASK]} token used in prompting, and entity marker tokens. Our method employs contrastive learning to extract complementary discriminative information from these individual representations. This is particularly relevant in low-resource settings where information is scarce. Leveraging multiple sentence representations is especially effective in distilling discriminative information for relation classification when additional information, like relation descriptions, are not available. We validate the adaptability of our approach, maintaining robust performance in scenarios that include relation descriptions, and showcasing its flexibility to adapt to different resource constraints\footnote{Our model is available at \url{https://github.com/pnborchert/MultiRep}.}.
\end{abstract}

\section{Introduction}

Relation classification (RC) is an important subtask in the relation extraction, focusing on identifying the types of relations between pairs of entities within a given textual context. 
Extracting relevant information is central to this task. To achieve this, RC models must distill rich information from sentences, including contextual cues, entity attributes, and relation characteristics. While language models are essential for extracting text representations, their use of vector space in sentence representations is suboptimal~\citep{ethayarajh-2019-contextual}. To improve this, recent advances have enhanced sentence representations through various techniques like flow-based methods~\citep{li-etal-2020-sentence}, whitening operations~\citep{huang-etal-2021-whiteningbert-easy}, prompting~\citep{jiang-etal-2022-promptbert}, and contrastive learning~\citep{gao-etal-2021-simcse,kim-etal-2021-self,zhou-etal-2022-debiased}.

\begin{figure*}[h]
    \vspace{-15pt}
    \centering
    \includegraphics[width=1\linewidth]{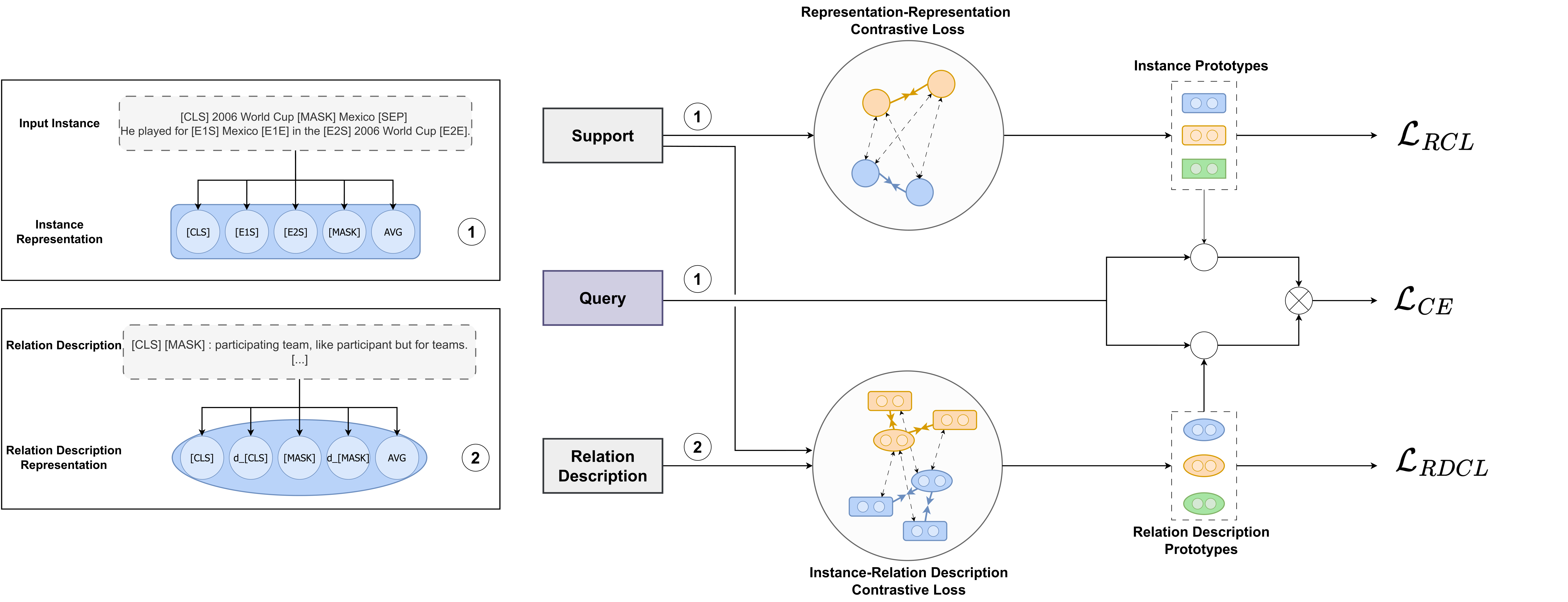}
    \caption{Overview of the MultiRep model, which integrates relation description information. The $\circ$ represents the vector dot product between the prototypes and the query samples, while the addition operation is denoted by $\otimes$. Attracting and repelling forces in contrastive learning are represented by $\rightarrow \leftarrow$ and $\dashleftarrow \dashrightarrow$, respectively. \circled{1} and \circled{2} illustrate the representations extracted from input sentences and relation descriptions, respectively.}
    \label{fig:model}
    \vspace{-10pt}
\end{figure*}

Relation extraction faces challenges with limited data for many relation types and disproportionate data acquisition costs~\cite{yang-etal-2021-entity}. To address this challenge, few-shot RC trains models to quickly adapt to new relation types using only few labeled examples. Common strategies include meta-learning and prototypical networks, leveraging representation similarity to match unseen query instances with few labeled support instances~\citep{snell_prototypical_2017}. Recent studies have enhanced model representations using supplementary data. \citet{yang-etal-2021-entity} and \citet{qu_fsre_bayesian} incorporate information from external knowledge bases, augmenting entity-related knowledge. \citet{wang-etal-2020-learning} and \citet{yu-etal-2022-dependency} utilize linguistic dependencies to integrate structural sentence information in models. Textual relation descriptions also improve prototypical network performance by offering additional insights into relation types~\citep{han-etal-2021-exploring, dong-etal-2021-mapre,liu-etal-2022-simple}.

Given the inherent complexity of distinguishing between various relation types, RC applications commonly combine representations of entity marker tokens as sentence representations~\citep{baldini-soares-etal-2019-matching,dong-etal-2021-mapre}. Recent work employs contrastive learning in few-shot RC for more discriminative representations~\cite{han-etal-2021-exploring,zhang-lu-2022-better,dong-etal-2021-mapre}. 
Additionally, studies show that representing sentences with the \texttt{[MASK]} token through prompting improves sentence representations \citep{jiang-etal-2022-promptbert}.

In this study, our goal is to enhance sentence representations for few-shot relation classification by aligning multiple representations, such as entity markers and the \texttt{[MASK]} token used in prompting. We recognize that individual model representations, being condensed summaries of the model's internal workings, often carry information that is not directly relevant to relation classification. To address this, we combine various representations to create more comprehensive sentence embeddings, thereby improving the model's discriminative power in low resource settings.
While multiple representations offer diverse perspectives, their informational content tends to overlap significantly. To align this information, we introduce contrastive learning objectives providing an effective solution for extracting complementary and discriminative information. A key advantage in our approach is the efficient utilization of resources, since all representations are derived from a single forward pass. Our method is adaptable to different resource constraints by varying the number of representations used in the sentence embeddings. This adaptability allows for a balance between performance and resource usage. Additionally, we extend our approach to include additional information sources, such as relation descriptions. In summary, our contributions are:
\begin{itemize}[leftmargin=*]
    \setlength{\itemsep}{0pt}
    \item We introduce a novel method for aligning multiple representations in few-shot relation classification using contrastive learning.
    \item Our approach adapts to various resource constraints and extends to additional information sources, including relation descriptions.
    \item We emphasize the resource efficiency of our method, improving performance in low-resource settings.
\end{itemize}

\section{Approach}

This section provides a detailed overview of our approach, as depicted in Figure \ref{fig:model}. 

\subsection{Sentence Representations} \label{sec:sentencerepresentations}

In line with related work, we utilize the BERT-Base model \citep{devlin-etal-2019-bert} to encode textual inputs \citep{han-etal-2021-exploring, liu-etal-2022-simple}. We describe below the techniques used to generate various sentence representations from the BERT encoder.

\textbf{Average Pooling} computes sentence representations by averaging token representations. Concurrently, the \texttt{[CLS]} token is used as a sentence representation during BERT-Base encoder's pretraining, capturing information of the entire input sequence.  The \textbf{entity marker} technique augments the input sentence $x$ with markers indicating entities in the text \citep{baldini-soares-etal-2019-matching}. This augments the input into $\bar x = [x_0, ..., \verb|[E1S]|, x_i, \verb|[E1E]|, ..., x_n]$. The sentence representation is constructed by concatenating the entity start marker representations $\verb|[E1S]|$ and $\verb|[E2S]|$ \citep{baldini-soares-etal-2019-matching}. In the \textbf{prompting} approach, the RC task is reformulated as a masked language modeling problem. With a template $\mathcal{T}$, each input is transformed into $x_{\text{prompt}}=\mathcal{T}(x)$ containing at least one \textbf{\texttt{[MASK]}} token. This masked token represents the relation label and is predicted from the context, e.g., $\bar x = \verb|[MASK]| \text{: } x \text{. }$ \citep{schick-schutze-2021-exploiting}. \citet{gao-etal-2021-simcse} introduce using \textbf{dropout} masks for generating augmented sentence representations. Since entity marker representations are not suitable for relation descriptions, we employ prompting and \texttt{[CLS]} representations with varying dropout masks.

\subsection{Contrastive Representation Learning}

The objective of our \textbf{representation-representation contrastive loss} term is aligning individual representations to extract discriminatory information for the relation classification task. A key difference to contrastive learning objectives in related work lies in our method of constructing positive instance pairs. Related methods like SimCSE \citep{gao-etal-2021-simcse}  or CMC \citep{tian2020contrastive} derive their representations from different augmented inputs. In contrast, we extract $M$ different representations from each input sentence in a single forward pass, and consider these representations as positive pairs. Consequently, representations from other sentences in the training set serve as negative instance pairs. For a given representation $r_{i}^{m}$ (where $m \in M$, $i \in N \times K$), we define positive instances $r_i^+$ and negative instances $r_i^-$ as follows:  
\vspace{-5pt} \begin{equation*}
\begin{split}
    & r_i^+ = \{ r_i^{k \neq m} \, | \, k \in M \} \\
    & r_i^- = \{ r_{j \neq i}^m \, | \, j \in N \times K \}
\end{split}
\end{equation*}

This aims to maximize the similarity between different representations of the same sentence and minimize the similarity to representations obtained from other sentences~\citep{oord2019representation,gao-etal-2021-simcse}.
It ensures that the differentiating factors encoded in the embeddings primarily reflect the underlying sentences, regardless of how these representations are derived from the internal model representations. 
The representation-representation contrastive loss is computed as follows:
\vspace{-5pt} \begin{equation*}
    \mathcal{L}_{RCL} = \sum_{i=1}^{N \times K}\sum_{m=1}^{M}{-log\frac{exp\left(\phi(r_i^m, r_i^+)/\tau \right)}{exp\left(\phi(r_i^m,r_j^-)/\tau\right)}},
\end{equation*}
where $\tau$ is a temperature scaling parameter, and $\phi(r_i^m,r_i^+)$ represents the element-wise cosine similarity $\sum_{k=1}^{M-1}{r_i^m \cdot r_{i}^k / \|r_i^m\| \|r_{i}^k\|}$ between representation $r_i^m$ and each representation in $r_i^+$.

In the \textbf{instance-relation description contrastive loss}, we leverage the relation descriptions to maximize the similarity between instance representations and corresponding relation description representations. To construct the instance representations $R_i$ and the relation description representations $D_i$, we concatenate all representations extracted from the encoder $R_i = [r_{i}^{1};r_{i}^{2};...;r_{i}^{M}]$ and $D_i = [d_{i}^{1};d_{i}^{2};...;d_{i}^{M}]$. Specifically, $R_i$ is composed of average pooled, \texttt{[CLS]}, \texttt{[MASK]}, and two entity marker representations. Given that entity marker representations are not available for relation descriptions, we apply a 10\% dropout to the \texttt{[CLS]} and \texttt{[MASK]} representations, along with the average pooled, \texttt{[CLS]}, and \texttt{[MASK]} representations to form $D_i$. For instance representation $R_i$, we select the corresponding relation description $D^+$ based on the label information in the support set. Non-corresponding relation descriptions $D^-$ form negative pairs. The instance-relation description contrastive loss is computed as follows:
\begin{equation*}
    \mathcal{L}_{RDCL} = \sum_{i=1}^{N \times K}{-log\frac{exp\left(\phi(R_i, D^+)/\tau \right)}{exp\left(\phi(R_i,D^-)/\tau\right)}}
\end{equation*}

\begin{table*}
\centering
\scalebox{0.8}{
    \begin{tabular}{lcrrrr|r}
    \hline
    \textbf{Model} & \textbf{Relation Descriptions}  & \textbf{5-1} & \textbf{5-5} & \textbf{10-1} & \textbf{10-5} & \textbf{Avg.} \\
    \hline

    Proto & - & - / 80.68 & - / 89.60 & - / 71.48 & - / 82.89 & - / 81.16 \\
    BERT-Pair & - & \underline{85.66} / 88.32 & 89.48 / 93.22 & 76.84 / 80.63 & 81.76 / 87.02 & 83.44 / 87.30 \\
    CTEG & - & 84.72 / 88.11 & \underline{92.52} / \textbf{95.25} & 76.01 / 81.29 & \underline{84.89} / \underline{91.33} & \underline{84.54} / 89.00\\
    DAPL & - & - / 85.94 & - / 94.28 & - / 77.59 & - / 89.26 & - / 86.77\\
    SimpleFSRE & - & 84.77 / \textbf{89.33} & 89.54 / 94.13 & \underline{76.85} / \underline{83.41} & 83.42 / 90.25 & 83.64 / \underline{89.28}  \\
    \textbf{MultiRep (Ours)} & - & \textbf{87.13} / \underline{89.20} & \textbf{92.93} / \underline{95.09} & \textbf{78.42} / \textbf{84.18} & \textbf{87.29} / \textbf{91.65} & \textbf{86.44} / \textbf{90.03} \\
    \hline
    TD-Proto & \checkmark & - / 84.76 & - / 92.38 & - / 74.32 & - / 85.92 & - / 84.34 \\
    HCRP & \checkmark & 90.90 / 93.76 & 93.22 / 95.66 & 84.11 / 89.95 & 87.79 / \underline{92.10} & 89.01 / 92.87\\
    SimpleFSRE & \checkmark & \underline{91.29} / \textbf{94.42} & \textbf{94.05} / \textbf{96.37} & \underline{86.09} / \underline{90.73} & \textbf{89.68} / \textbf{93.47} & \underline{90.28} / \textbf{93.75} \\
    \textbf{MultiRep (Ours)} & \checkmark & \textbf{92.73} / \underline{94.18} & \underline{93.79} / \underline{96.29} & \textbf{86.12} / \textbf{91.07} & \underline{88.80} / 91.98 & \textbf{90.36} / \underline{93.38} \\

    \hline
    \end{tabular}}
    \caption{Accuracy on the FewRel validation / test set.}
    \label{tab:results}
    \vspace{-10pt}
\end{table*}
\begin{table}
\centering
\scalebox{0.9}{
    \begin{tabular}{lrr}
    \hline
    \textbf{Model}  & \textbf{5-1} & \textbf{10-1} \\
    \hline

    MultiRep & 92.73 & 86.12 \\
    \hline
    w/o $\mathcal{L}_{RCL}$ & 92.08 & 85.95 \\
    w/o $\mathcal{L}_{RDCL}$ & 90.14 & 84.16 \\
    \hline
    w/o Avg. Pooling & 92.26 & 85.82 \\ 
    w/o Entity Marker & 91.90 & 84.83 \\ 
    w/o [CLS] & 91.35 & 85.51 \\ 
    w/o [MASK] & 91.87 & 85.80\\ 
    \hline
    w/ prototype addition  & 91.75 & 85.82 \\

    \hline
    \end{tabular}}
    \caption{Model variants with (w/) or without (w/o) indicated representations and architectural changes evaluated on the FewRel validation set.}
    \label{tab:ablations}
\end{table}

\subsection{Relation Classification}

We obtain $N$ class prototypes by averaging the $K$ instance representations in the support set. We compute the similarity between query instances and support prototypes using the vector dot product and selecting the most similar class prototype. 
For the relation description, we compute the similarity between query instances and relation description representations $D$. We add the similarities obtained from the relation descriptions with the similarities obtained from the class prototypes and select the most similar prototype and relation description. This is in line with \citet{liu-etal-2022-simple}, who instead directly add the prototype and relation description representations. We compute the cross-entropy loss $\mathcal{L}_{CE}=-log\left( z_y \right)$, where $z_y$ is the probability for class $y$. The total loss is defined as the sum of the individual loss terms $\mathcal{L} = \mathcal{L}_{CE} + \mathcal{L}_{RCL} + \mathcal{L}_{RDCL}$.

\section{Experiments}

\subsection{Task Definition}

In the N-way K-shot evaluation setting, episodes are randomly sampled from the training set. An episode consists of $N \times K$ input sentences $x$ in the support set $S=\{(x_{i}, rel_{i})\}^{N \times K}_{i=1}$ and $N \times K$ inputs from the query set $Q=\{x_i\}^{N \times K}_{i=1}$. The relations are randomly sampled from the relation types included in the training dataset. Importantly, the relation types in the training set are not overlapping with the test set (and validation set) $rel\textsubscript{train} \cap rel\textsubscript{test} = \varnothing$ \citep{gao-etal-2019-fewrel}.

\subsection{Dataset and Evaluation}

Our experiments were conducted using the FewRel dataset, which consists of 700 instances for each of the 100 relation types \citep{han-etal-2018-fewrel}. Originating from Wikipedia, the dataset is segmented into training, validation, and test sets, containing 64, 16, and 20 relation types, respectively.
Following the training and evaluation procedure outlined in \citet{gao-etal-2019-fewrel}, we train MultiRep on randomly sampled episodes from the training set and subsequently evaluate the model on previously unseen data from the validation and test sets. We report classification accuracy, with validation accuracy metrics being averaged over three random seeds. The test set accuracies are referenced from the public FewRel leaderboard\footnote{Our evaluation results on the FewRel test set are available on the public leaderboard, listed under the ``multirep'' alias.}. Our MultiRep model is trained for 30,000 iterations on the FewRel training set with a batch size of 4 and a learning rate of 2e-5, ensuring our experiments are consistent with related work \citep{han-etal-2021-exploring, liu-etal-2022-simple}.

\subsection{Results}

We present the results of our MultiRep approach and compare them to relevant benchmark models designed for few-shot RC, some of which incorporate relation descriptions as additional information. For a consistent and fair model comparison, all benchmarked models exclusively utilize BERT-Base \citep{devlin-etal-2019-bert} as the sentence encoder. This approach aligns with recent findings suggesting that LLMs like GPT-3.5 do not yield superior performance in relation extraction tasks \citep{meng-etal-2023-rapl}. The benchmarked models include Proto \citep{gao-etal-2019-fewrel}, BERT-Pair \citep{gao-etal-2019-fewrel}, TD-Proto \citep{yang2020_enhance}, CTEG \citep{wang-etal-2020-learning-decouple}, DAPL \citep{yu-etal-2022-dependency}, HCRP \citep{han-etal-2021-exploring}, and SimpleFSRE \citep{liu-etal-2022-simple}.

Our model evaluation results are summarized in Table \ref{tab:results}. We analyze these results for two distinct scenarios: (i) models that do not incorporate additional information, and (ii) models that incorporate relation description information. 
We observe that MultiRep outperforms existing models, particularly in settings where information is limited. Specifically, this includes scenarios where relation description information is unavailable, as well as 1-Shot settings in the presence of relation description information. 
To validate the importance of individual components in the MultiRep model, we conducted ablation studies and present the results in Table \ref{tab:ablations}. These results are based on the MultiRep model that incorporates relation description information, evaluated on the FewRel validation set in the 5-Way 1-Shot and 10-Way 1-Shot settings. Our findings indicate that removing the contrastive learning loss terms, $\mathcal{L}_{RCL}$ and $\mathcal{L}_{RDCL}$, substantially reduces model performance. Furthermore, removing individual representations from the MultiRep model has a negative impact on performance, and there are no specific representations that disproportionately affect the model's performance. Additionally, we validate our approach of computing separate instance prototypes and relation description prototypes, as compared to the direct prototype addition method introduced by \citet{liu-etal-2022-simple}. 

\subsection{Resource Adaptability}

\begin{figure}
    \centering
    \includegraphics[width=1\linewidth]{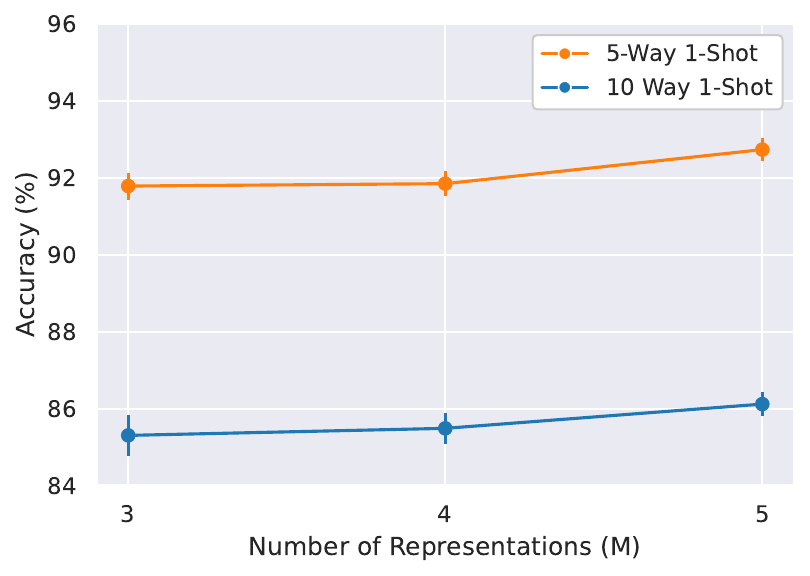}
    \caption{Average accuracy and standard deviation for varying number of representations $M$ evaluated on the FewRel validation set.}
    \label{fig:ablation_representations}
    \vspace{-5pt}
\end{figure}

To assess our method's adaptability to different resources, we explored how changes in the number of representations affect relation classification performance. We experimented with different combinations of individual representations as outlined in Section \ref{sec:sentencerepresentations}. Figure \ref{fig:ablation_representations} shows the average accuracy obtained with differing numbers of representations ($M$) on the FewRel validation set, employing our MultiRep model that includes relation descriptions. The analysis revealed a clear trend: increasing the number of representations leads to improved performance and reduced performance variability. The results demonstrate that the MultiRep approach optimizes model performance and indicate that the number of representations can be adjusted flexibly to accommodate resource constraints.

\section{Conclusion}

In this study, we propose aligning multiple sentence representations in few-shot RC through contrastive learning. This method efficiently distills discriminative information from these representations. Our results emphasize its adaptability and effectiveness in diverse scenarios, particularly in low-resource and 1-Shot environments, particularly when relation descriptions are unavailable. 
A key advantage of our approach lies in its resource efficiency, deriving all sentence representations from a single forward pass. 
Additionally, we demonstrate that the performance of our method improves with an increasing number of representations, showcasing its resource-adaptability.

\section{Limitations}

Although our approach efficiently utilizes multiple sentence representations within a single forward pass, it is important to note that this involves combining these representations into larger vectors. This aggregation process may require additional memory and computational resources. Moreover, the application of contrastive learning comes with additional computational requirements. Our method is specifically designed for few-shot RC tasks, and its performance might vary when applied to different types of NLP tasks.

\bibliography{anthology,references}

\clearpage
\appendix

\section{Model Training}
\label{sec:appendix}

The MultiRep model consists of 109.48 million parameters and was trained on a single NVIDIA A6000 48GB GPU. The combined training and evaluation time for the 5-Way 1-Shot and 10-Way 5-Shot models, incorporating relation descriptions, was 16 hours and 25 hours, respectively.

\section{Input Templates}
\label{sec:appendix_templates}

To obtain various representations for a given input sequence and relation description within a single forward pass, we augment the inputs with special tokens. For input sequences, we employ the template ``\verb|[CLS]| \verb|[E1]|, \verb|[MASK]|, \verb|[E2]| \verb|[SEP]| \verb|[TEXT]|'', where \verb|[E1]| and \verb|[E2]| denote the respective entities in the input sentence, and \verb|[TEXT]| represents the augmented input sentence with entity markers. For the relation description, we utilize the template ``\verb|[CLS]| \verb|[MASK]|: \verb|[RELATION DESCRIPTION]|'', where the relation descriptions included in the FewRel dataset are represented by \verb|[RELATION DESCRIPTION]|. The FewRel dataset provides a relation type category and textual description of the relation type. We include both the relation type category and textual description, separated by a comma in the \verb|[RELATION DESCRIPTION]|.

\section{Case Study}

This case study highlights the contributions of our MultiRep approach, which integrates representation-representation contrastive learning objective. We demonstrate this by comparing the embeddings and predictions of the MultiRep model, trained with these contrastive learning objective, against a variant trained without them (MultiRep w/o CL). In a 5-Way 1-Shot setting on the FewRel validation set, the MultiRep w/o CL model achieves an accuracy of 84.83\%, substantially lower than the 87.13\% achieved by the MultiRep model with contrastive learning objectives.

\begin{figure}[!tbp]
  \centering
  \subfloat{\includegraphics[width=0.48\linewidth]{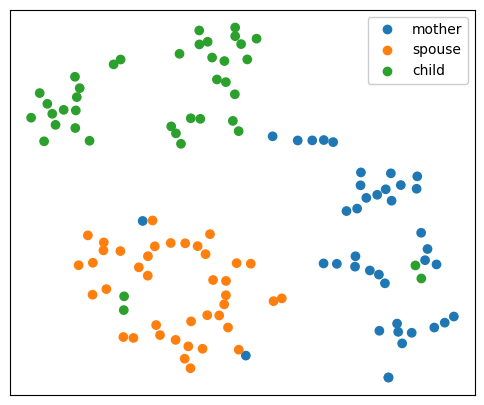}}
  \hfill
  \subfloat{\includegraphics[width=0.48\linewidth]{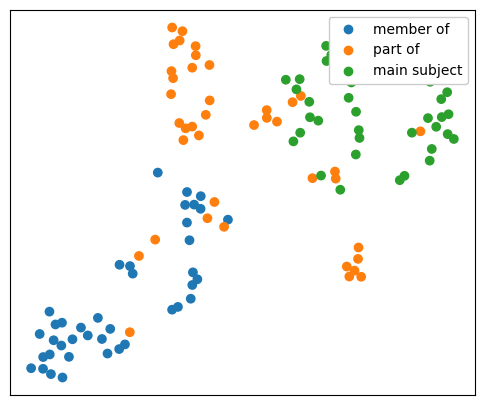}}
  \caption{t-SNE visualization of 120 randomly sampled support embeddings for hard relation classification examples obtained from MultiRep.}
  \label{fig:tsne_multirep}
\end{figure}

\begin{figure}[!tbp]
  \centering
  \subfloat{\includegraphics[width=0.48\linewidth]{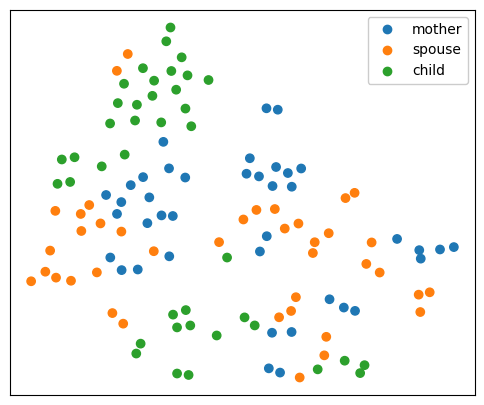}}
  \hfill
  \subfloat{\includegraphics[width=0.48\linewidth]{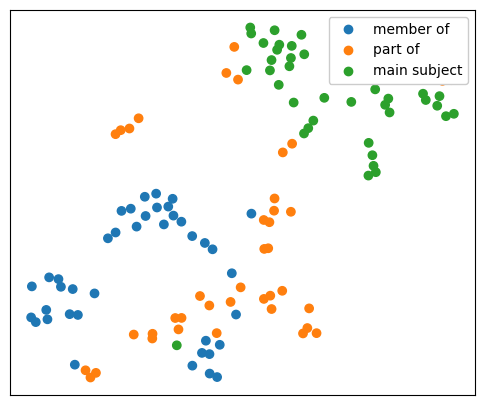}}
  \caption{t-SNE visualization of 120 randomly sampled support embeddings for hard relation classification examples obtained from MultiRep w/o CL.}
  \label{fig:tsne_nocl}
\end{figure}

Both models combine five individual representations to form sentence representations, as detailed in Section \ref{sec:sentencerepresentations}. We display selected prediction results in Table \ref{tab:prediction_examples}. Notably, our MultiRep model demonstrates a tendency to misclassify more challenging cases, such as instances involving difficult relation types like "spouse", "mother", and "child". This observation aligns with the hard classification examples highlighted and addressed by \citet{han-etal-2021-exploring}.
Conversely, the model variant without contrastive learning objectives is more prone to errors in simpler classification tasks, leading to an overall lower performance. To further demonstrate the benefit of our contrastive representation learning approach, we present a t-SNE projection of sentence representations from both the MultiRep and MultiRep w/o CL models in Figures \ref{fig:tsne_multirep} and \ref{fig:tsne_nocl}. These figures allow for a direct comparison of sentence embeddings for relation types that were frequently confused, as listed in Table \ref{tab:prediction_examples}. The projections reveal that sentence embeddings for the same relation type cluster more closely together, while those from different types are more dispersed. This clustering pattern suggests that the sentence embeddings obtained from our MultiRep model effectively extract discriminative information for relation classification.

\begin{table*}
\centering
\scalebox{0.9}{
    \begin{tabular}{l|p{2cm}|p{2cm}|p{10cm}}
    \hline
    \textbf{Label} & \textbf{Prediction MultiRep} & \textbf{Prediction MultiRep w/o CL} & \textbf{Sentence} \\
    \hline

    sport & sport & military rank & \entityOneFull{Jürgen Hasler \entityOneTitle{E1}} ( born 7 May 1973 ) is a Liechtenstein former \entityTwoFull{alpine skier \entityTwoTitle{E2}} who competed in the 1994 Winter Olympics , 1998 Winter Olympics and 2002 Winter Olympics. \\
    &&&\\
    main subject & main subject & constellation & Comet 252P / \entityOneFull{LINEAR \entityOneTitle{E1}} is a periodic comet and \entityTwoFull{near - Earth object \entityTwoTitle{E2}} discovered by the LINEAR survey on April 7 , 2000. \\
    &&&\\
    part of & part of & competition class & During the \entityOneFull{Caucasus Campaign \entityOneTitle{E1}} of \entityTwoFull{World War I \entityTwoTitle{E2}}, the Russian Empire occupied most of the Armenian - populated regions of the Ottoman Empire. \\
    &&&\\
    spouse & spouse & mother & He was the father - in - law of Azerbaijan 's President \entityTwoFull{Heydar Aliyev \entityTwoTitle{E2}}, who married his daughter \entityOneFull{Zarifa Aliyeva \entityOneTitle{E1}}, and maternal grandfather of Azerbaijan 's current President Ilham Aliyev. \\
    &&&\\
    child & child & mother & He was the father of \entityOneFull{Frank Islacker \entityOneTitle{E1}} and the grandfather of \entityTwoFull{Mandy Islacker \entityTwoTitle{E2}}. \\
    &&&\\

    \hline

    child & mother & child & "Me and Liza" is about Wainwright's relationship with \entityTwoFull{Liza Minnelli \entityTwoTitle{E2}}, who was reportedly upset by his 2006 tribute concerts to her mother , American actress and singer \entityOneFull{Judy Garland \entityOneTitle{E1}}. \\

    &&&\\
    
    spouse & mother & spouse & His brother , Fridolin Weber , who died in 1785 , was the father of \entityTwoFull{Mozart \entityTwoTitle{E2}} 's wife , \entityOneFull{Constanze \entityOneTitle{E1}}. \\

    &&&\\
    
    member of & part of & military rank & Designers \entityOneFull{Domenico Dolce \entityOneTitle{E1}} and Stefano Gabbana ( \entityTwoFull{Dolce \& Gabbana \entityTwoTitle{E2} } ) spoke of working with Minogue for the costumes of the tour . \\
    &&&\\
    main subject & member of & part of & His grandmother Elisabeth was a member of the \entityTwoFull{Ephrussi family \entityTwoTitle{E2}}, whose history he chronicled in " \entityOneFull{The Hare with Amber Eyes \entityOneTitle{E1} } ". \\
    &&&\\
    part of & constellation & constellation & The SS " Edward Y. Townsend " ( official number 203449 ) was a \entityOneFull{American \entityOneTitle{E1}} Great Lakes freighter that served on the Great Lakes of \entityTwoFull{North America \entityTwoTitle{E2}}. \\

    \hline
    \end{tabular}}
    \caption{Comparative prediction examples: MultiRep with contrastive learning objectives (MultiRep) versus MultiRep without contrastive learning objectives (w/o CL). Examples are obtained from the FewRel validation set.}
    \label{tab:prediction_examples}
\end{table*}

\end{document}